\documentclass{bmvc2k}

\title{Few-Shot Viewpoint Estimation}

\addauthor{Hung-Yu Tseng}{htseng6@ucmerced.edu}{1}
\addauthor{Shalini De Mello}{shalinig@nvidia.com}{2}
\addauthor{Jonathan Tremblay}{jtremblay@nvidia.com}{2}
\addauthor{Sifei Liu}{sifeil@nvidia.com}{2}
\addauthor{Stan Birchfield}{sbirchfield@nvidia.com}{2}
\addauthor{Ming-Hsuan Yang}{mhyang@ucmerced.edu}{1}
\addauthor{Jan Kautz}{jkautz@nvidia.com}{2}

\addinstitution{
 University of California, Merced
}
\addinstitution{
 NVIDIA
}

\runninghead{Tseng \etal}{Few-Shot Viewpoint Estimation}

\def\eg{\emph{e.g}\bmvaOneDot}

\def\etal{\emph{et al}\bmvaOneDot}

\usepackage{xspace}
\usepackage{color}
\usepackage{multirow}
\usepackage{ifthen}
\usepackage{algorithmicx}
\usepackage{algorithm}
\usepackage{algpseudocode}
\usepackage{colortbl}  

\graphicspath{{figures/}}

\newcommand{\tb}[1]{\textbf{#1}}

\makeatletter
\newcommand\semitiny{\@setfontsize\notsotiny{6.31415}{7.1828}}
\makeatother

\long\def\ignorethis#1{}

\definecolor{lightgray}{rgb}{0.92,0.92,0.92}
\definecolor{gray}{rgb}{0.35,0.35,0.35}
\definecolor{orange}{rgb}{0.5,0.5,0}
\definecolor{MyBlue}{rgb}{0,0.2,0.8}
\definecolor{MyRed}{rgb}{0.8,0.2,0}
\definecolor{MyGreen}{rgb}{0.0,0.5,0.1}
\definecolor{MyGray}{rgb}{0.4,0.4,0.4}
\definecolor{airforceblue}{rgb}{0.36, 0.54, 0.66}

\newlength\paramargin
\newlength\figmarginstart
\newlength\figmargin
\newlength\subfigmargin
\newlength\secmargin
\newlength\subsecmargin
\newlength\tabmargin
\newlength\tabmarginstart
\newlength\eqmargin
\newlength\algmargin

\usepackage{array}
\newcolumntype{L}[1]{>{\raggedright\let\newline\\\arraybackslash\hspace{0pt}}m{#1}}
\newcolumntype{C}[1]{>{\centering\let\newline\\\arraybackslash\hspace{0pt}}m{#1}}
\newcolumntype{R}[1]{>{\raggedleft\let\newline\\\arraybackslash\hspace{0pt}}m{#1}}

\newcommand{\Paragraph}[1]
{\vspace{1.5mm} \noindent \textbf{#1}}

\def\eg{e.g.,~}
\def\etc{etc.}

\def\etal{et~al.\xspace}

\newcommand{\secref}[1]{Section~\ref{sec:#1}}

\newcommand{\figref}[1]{Figure~\ref{figure:#1}}
\newcommand{\tabref}[1]{Table~\ref{tab:#1}}
\newcommand{\eqnref}[1]{\eqref{eq:#1}}

\usepackage{epsfig}
\usepackage{graphicx}
\usepackage{amsmath}
\usepackage{amssymb}
\usepackage{paralist}
\usepackage{booktabs}
\usepackage{gensymb}
\usepackage[title]{appendix}

\begin{document}

\maketitle
\begin{figure}[th]
	\centering
	\vspace{-3mm}
    \includegraphics[width=0.95\linewidth]{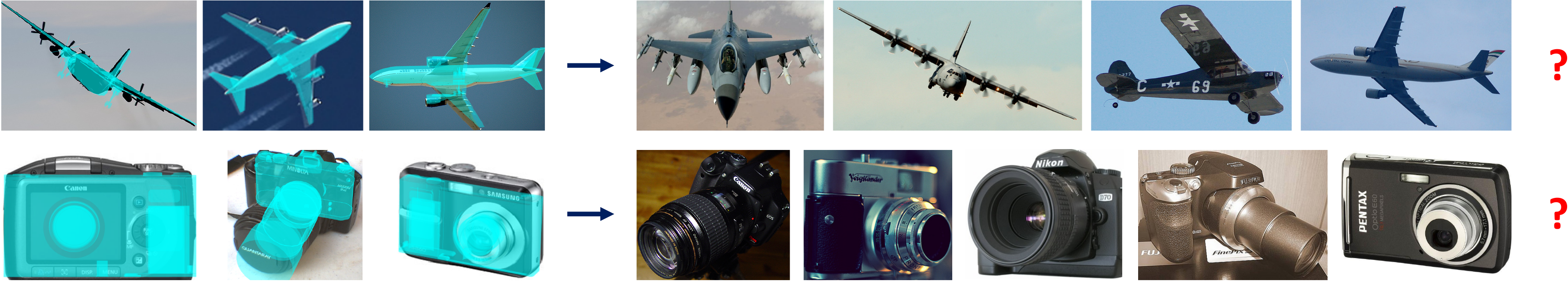}
    \caption{\textbf{Few-shot viewpoint estimation.} Given only a \textit{few} images of a novel category with annotated viewpoints (left images with rendered CAD models{\protect\footnotemark}), we aim to learn to predict the viewpoint of arbitrary objects from the same category.}
    \vspace{-1mm}
    \label{fig:problem}
\end{figure}

\begin{abstract}
Viewpoint estimation for known categories of objects has been improved significantly thanks to deep networks and large datasets, but generalization to \emph{unknown} categories is still very challenging. With an aim towards improving performance on unknown categories, we introduce the problem of category-level few-shot viewpoint estimation. We design a novel framework to successfully train viewpoint networks for new categories with few examples ($10$ or less). 
We formulate the problem as one of learning to estimate category-specific 3D canonical shapes, their associated depth estimates, and semantic 2D keypoints.
We apply meta-learning to learn weights for our network that are amenable to category-specific few-shot fine-tuning.
Furthermore, we design a flexible meta-Siamese network that maximizes information sharing during meta-learning.
Through extensive experimentation on the  ObjectNet3D and Pascal3D+ benchmark datasets, we demonstrate that our framework, which we call MetaView, significantly outperforms fine-tuning the state-of-the-art models with few examples, and that the specific architectural innovations of our method are crucial to achieving good performance.
\end{abstract}


\vspace{\secmargin}
\section{Introduction}
\vspace{\secmargin}

Estimating the viewpoint (azimuth, elevation, and cyclorotation) of rigid objects, relative to the camera, is a fundamental problem in three-dimensional (3D) computer vision. It is vital to applications such as robotics~\cite{tremblay2018synthetically}, 3D model retrieval~\cite{grabner20183d}, and reconstruction~\cite{kundu20183d}. With convolutional neural networks~(CNNs) and the availability of many labeled examples~\cite{chang2015shapenet, xiang2016objectnet3d, xiang_pascal3d}, much progress has been made in estimating the viewpoint of \emph{known} categories of objects~\cite{grabner20183d, mousavian20173d, pavlakos20176}. 
However, it remains challenging for even the best methods~\cite{zhou2018starmap} to generalize well to \emph{unknown} categories that the system has not encountered during training~\cite{kuznetsova2016exploiting, tulsiani2015pose, zhou2018starmap}. In such a case, re-training the viewpoint estimation network on an unknown category would require annotating thousands of new examples, which is labor-intensive.
\footnotetext{We do not use the CAD models in our method, and we show them here for the purpose of illustrating viewpoint.} 

To improve the performance of viewpoint estimation on unknown categories with little annotation effort, we introduce the problem of \emph{few-shot viewpoint estimation}, in which a few ($10$ or less) labeled training examples are used to train a viewpoint estimation network for each novel category. 
We are inspired by the facts that (a) humans are able to perform mental rotations of objects~\cite{shepard1971mental} and can successfully learn novel views from a few examples~\cite{palmer1999vision}; and (b) recently, successful few-shot learning methods for several other vision tasks have been proposed~\cite{finn2017maml, gui2018few, park2018meta}.

However, merely fine-tuning a viewpoint estimation network with a few examples of a new category can easily lead to over-fitting.
To overcome this problem, we formulate the viewpoint estimation problem as one of learning to estimate category-specific 3D canonical keypoints, their 2D projections, and associated depth values from which viewpoint can be estimated.
We use meta-learning~\cite{andrychowicz2016learning, finn2017maml} to learn weights for our viewpoint network that are optimal for category-specific few-shot learning.
Furthermore, we propose meta-Siamese, which is a flexible network design that maximizes information sharing during meta-learning and adapts to an arbitrary number of keypoints. 
Through extensive evaluation on the ObjectNet3D~\cite{xiang2016objectnet3d} and Pascal3D+~\cite{xiang_pascal3d} benchmark datasets, we show that our proposed method helps to significantly improve performance on unknown categories and outperforms fine-tuning the state-of-the-art models with a few examples of new categories.

To summarize, the main scientific contributions of our work are:
\begin{compactitem}

\item 
We introduce the problem of category-level few-shot viewpoint estimation, thus bridging viewpoint estimation and few-shot learning.

\item 
We design a novel meta-Siamese architecture and adapt meta-learning to learn weights for it that are optimal for category-level few-shot learning.

\end{compactitem}
\vspace{-0.3cm}

\vspace{\secmargin}
\section{Related work}
\vspace{\secmargin}
\Paragraph{Viewpoint estimation.}
Many viewpoint estimation networks have been proposed for single~\cite{kundu20183d, Su_2015_ICCV, tulsiani2015viewpoints} or multiple~\cite{grabner20183d, zhou2018starmap} categories; or individual instances~\cite{rad2018feature, Sundermeyer2018eccv:implicit} of objects. They use different network architectures, including those that estimate angular values directly~\cite{kehl2017ssd, mousavian20173d, Su_2015_ICCV, tulsiani2015viewpoints, xiang2018rss:posecnn}; encode images in latent spaces to match them against a dictionary of ground truth viewpoints~\cite{massa2016deep, Sundermeyer2018eccv:implicit}; or detect projections of 3D bounding boxes~\cite{grabner20183d, rad2017bb8, tekin2017real, tremblay2018synthetically} or of semantic keypoints~\cite{pavlakos20176, zhou2018starmap}, which along with known~\cite{pavlakos20176} or estimated~\cite{zhou2018starmap, grabner20183d} 3D object structures are used to compute viewpoint. Zhou~\etal propose the state-of-the-art StarMap method that detects multiple visible general keypoints~\cite{zhou2018starmap} similar to SIFT~\cite{lowe2004distinctive} or SURF~\cite{bay2008surf} via a learned CNN, and estimates category-level canonical 3D shapes. The existing viewpoint estimation methods are designed for known object categories and hence very few works report performance on unknown ones~\cite{kuznetsova2016exploiting, tulsiani2015pose, zhou2018starmap}. Even highly successful techniques such as~\cite{zhou2018starmap} perform significantly worse on unknown categories versus known ones. To our knowledge, no prior work has explored few-shot learning as a means to improve performance on novel categories and our work is the first to do so.

The existing viewpoint estimation networks also require large training datasets and two of them: Pascal3D+~\cite{xiang_pascal3d} and ObjectNet3D~\cite{xiang2016objectnet3d} with 12 and 100 categories, respectively, have helped to move the field forward. At the instance level, the LineMOD~\cite{hinterstoisser2012accv:linemod}, T-LESS~\cite{hodan2017wacv:tless}, OPT~\cite{wu2017ismar:opt}, and YCB-Video~\cite{xiang2018rss:posecnn} datasets that contain images of no more than 30 known 3D objects are widely used. Manual annotation of object viewpoint by aligning 3D CAD models to images (\eg Figure~(1)); or of 2D keypoints is a significant undertaking. To overcome this limitation, viewpoint estimation methods based on unsupervised learning~\cite{suwajanakorn2018key-pointnet}; general keypoints~\cite{zhou2018starmap}; and synthetic images~\cite{rad2018feature, Su_2015_ICCV, Sundermeyer2018eccv:implicit, tremblay2018pose, xiang2018rss:posecnn} have been proposed.

\Paragraph{Few-shot learning.} Successful few-shot learning algorithms for several vision tasks, besides viewpoint estimation, have been proposed recently. These include object recognition~\cite{finn2017maml,ravi2017metalstm,rezende2016oneshot,santoro2016meta,snell2017prototypical,vinyals2016matching}, segmentation~\cite{rakelly2018few,shaban2017one}, online adaptation of trackers~\cite{park2018meta}, and human motion prediction~\cite{gui2018few}. Several of these methods use meta-learning~\cite{andrychowicz2016learning} to learn a ``learner'' that is amenable to few-shot learning of a specific task from a set of closely related tasks. The learner may take the form of (a) a training algorithm~\cite{finn2017maml,Nichol2018:reptile,ravi2017metalstm}; (b) a metric-space for representing tasks~\cite{snell2017prototypical,vinyals2016matching}; or (c) a meta-recurrent network~\cite{rezende2016oneshot,santoro2016meta}. The MAML~\cite{finn2017maml} meta-learning algorithm that learns a set of network initialization weights that are optimal for few-shot fine-tuning, is shown to be useful for many vision tasks.

Relative to the existing work, in this work we train networks for category-level viewpoint estimation. We further assume that we do not have access to 3D CAD models of any object or category. Lastly, we endeavor to train viewpoint networks for new categories with very few examples---a task that has not been attempted previously.
\vspace{-0.3cm}

\vspace{\secmargin}
\section{Few-shot Viewpoint Estimation}\label{sec:method}
\vspace{\secmargin}

\begin{figure*}[t]
	\centering
    \includegraphics[width=0.9\linewidth]{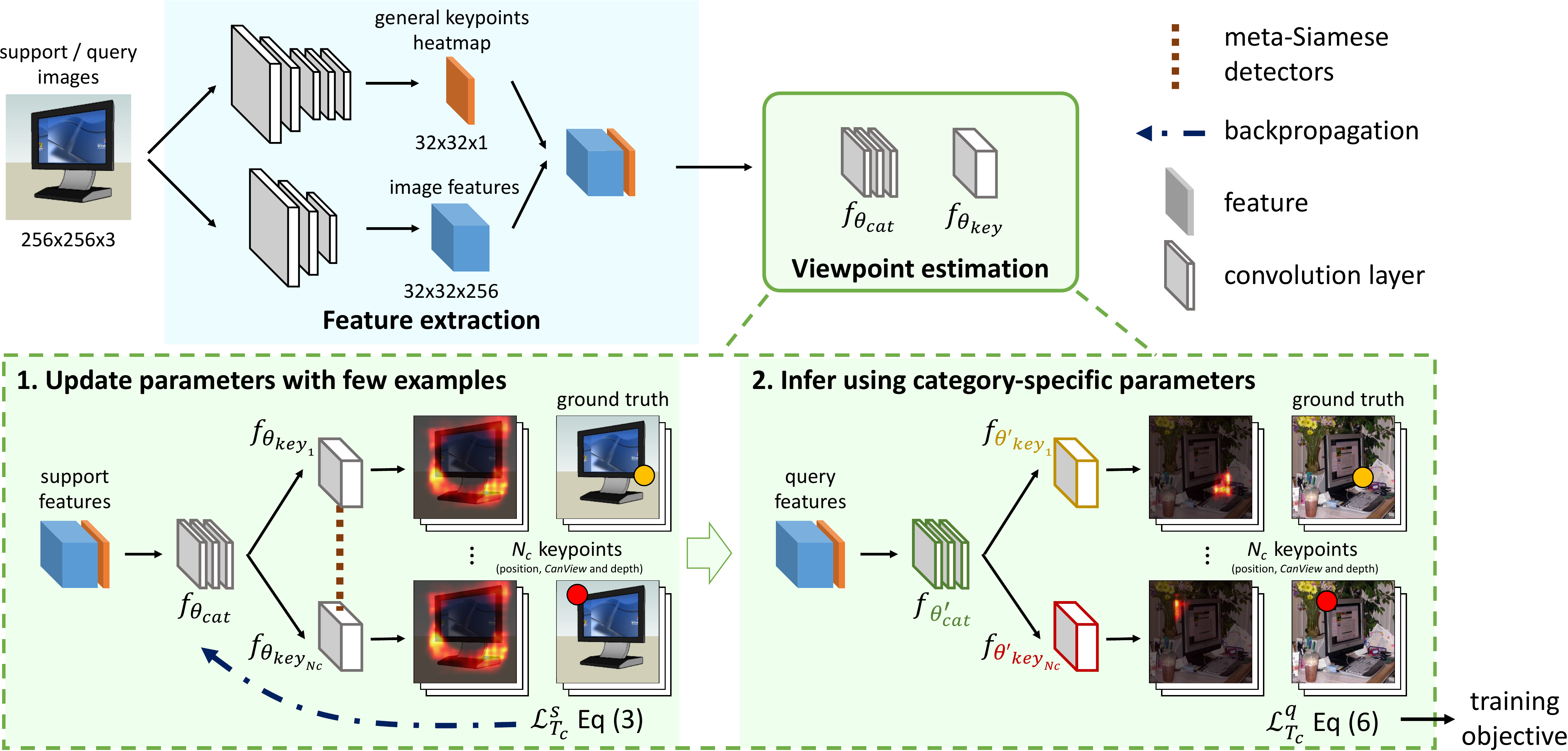}
    \vspace{-3mm}
    \caption{\textbf{Method overview.} 
    Our MetaView framework is composed of a category-agnostic feature extraction 
    (top-left) and category-specific viewpoint estimation (top-right) blocks.
    The bottom components show the different steps for training our viewpoint estimation block via meta-learning or for adapting it to a new category (bottom left only), which are described in detail in Section~\ref{sec:method}.
    }
    \vspace{\figmargin}
    \label{figure:overview}
\end{figure*}

Our proposed MetaView framework for category-level few-shot viewpoint estimation is shown in the top row of~\figref{overview}. It consists of two main components: a category-agnostic feature extraction block designed to extract general features from images that help to improve the accuracy of the downstream viewpoint estimation task; and a category-specific viewpoint estimation block designed to compute the viewpoint of all objects of a specific category. The latter block, in turn, computes viewpoint by detecting a unique set of semantic keypoints (containing 3D, 2D and depth values) via a category-specific feature extraction module ($f_{\theta_{\it cat}}$) and a category-specific keypoint detection module ($f_{\theta_{\it key}}$).

Our system operates in the following manner. We first train each of our feature extraction and viewpoint estimation blocks using a training set $S^\mathrm{train}$ containing a finite set of object categories. 
We use standard supervised learning to train the feature extraction block and fix its weights for all subsequent training stages.
We then use meta-learning to train our viewpoint estimation block. It uses an alternative training procedure designed to make the viewpoint estimation block an effective few-shot ``learner''. This means that when our trained viewpoint estimation block is further fine-tuned with a few examples of an unknown category, it can generalize well to other examples of that category.

At inference time, we assume that our system encounters a new category (not present during training) along with a few of its labeled examples from another set $S^\mathrm{test}$ (\eg the category ``monitor'' shown in the lower part of~\figref{overview}). We construct a unique viewpoint estimation network for it, initialize its weights with the optimal weights $\theta_{\it cat}^*$ and $\theta_{\it key}^*$ learned during meta-learning, and fine-tune it with the new category's few labeled examples (lower left of \figref{overview}). This results in a category-specific viewpoint network that generalizes well to other examples of this new category (lower right of \figref{overview}). In the following sections, we describe the architecture and the training procedure of each component in more detail.

\vspace{\secmargin}
\subsection{Feature Extraction}
\vspace{\subsecmargin}
\label{sec:3_2}

The first stage of our pipeline is a feature extraction block (top left of~\figref{overview}), which we train and use to extract features without regard to an object's category. It consists of two ResNet-18-style~\cite{he2016deep} networks: one trained as described in~\cite{zhou2018starmap} to extract a multi-peak heatmap for the locations of many visible general keypoints (see examples in the supplementary material); and another whose first four convolutional blocks compute an identically-sized set of high-level convolutional features and is trained to detect 8 semantic keypoints for all categories by optimizing the loss in Eq.~(\ref{eq:queryloss}) described later in Section~\ref{sec:metaTrain}. We concatenate the multi-peak heatmap and high-level features and input them to the viewpoint estimation block. We train the feature extraction block via standard supervised SGD learning and once trained, we fix its weights for all subsequent steps.

\vspace{\secmargin}
\subsection{Viewpoint Estimation}

Our viewpoint estimation block (top right in \figref{overview}) is specific to each category. It computes a 3D canonical shape for each category, along with its 2D image projection and depth values; and relates these quantities to compute an object's viewpoint. Furthermore, it is trained via meta-learning to be an optimal few-shot ``learner'' for any new category. We describe its architecture and training procedure in the following sections.

\vspace{\secmargin}
\subsubsection{Architecture}
\vspace{\secmargin}

\Paragraph{Viewpoint estimation via semantic keypoints.}
We assume that we have no knowledge of the 3D shape of any object in a category. So, to compute viewpoint, inspired by~\cite{zhou2018starmap}, we train our viewpoint estimation block to estimate a set of 3D points $\{(x_k, y_k, z_k) | k=1 \ldots N_c\}$, which together represent a canonical shape for the entire category $\mathcal{T}_{c}$ in an object-centric coordinate system (\eg for the category ``chairs'' it may comprise of the corners of a stick-figure representation of a prototypical chair with a back, a seat, and 4 legs). Additionally, for each 3D point $k$, our network detects its 2D image projection $(u_k, v_k)$ and estimates its associated depth $d_k$. We refer collectively to the values $(x_k, y_k, z_k
)$, $\left(u_k, v_k\right)$, $d_k$ of a point $k$ as a ``\emph{semantic keypoint}''. Finally, we obtain the viewpoint (rotation) of an object by solving the set of equations that relate each of the $k$ rotated and projected 3D canonical points $(x_k, y_k, z_k)$ to its 2D image location and depth estimate $(u_k, v_k, d_k)$, via orthogonal Procrustes. Note that our viewpoint estimation block is different from that of Zhou~\etal's~\cite{zhou2018starmap} as they detect the 2D projections of only the \emph{visible} 3D canonical points, whereas we detect projections of \emph{all} visible and invisible ones, thus providing more data for estimating viewpoint.

\Paragraph{Semantic keypoint estimation.} To locate the 2D image projection $(u_k,v_k)$ of each 3D keypoint $k$, the output of our network is a 2D heatmap $h_k(u, v)$, which predicts the probability of the point being located at $(u, v)$. It is produced by a spatial softmax layer. We obtain the final image coordinates $(u_k,v_k)$ via a weighted sum of the row $(u)$ and column $(v)$ values as:
\begin{equation}
\begin{aligned}
 (u_k, v_k)=\sum_{u, v} (u, v) \cdot h_k(u, v).
\end{aligned}.
\vspace{\eqmargin}
\end{equation}

Our network similarly computes a 2D map of depth values $c_k(u, v)$ that is of the same size as $h_k(u, v)$, along with three more maps $m_k^{i = \{ x, y, z \}}(u, v)$ for each dimension of its 3D canonical keypoint. The final depth estimate $d_k$ and 3D keypoint $(x_k, y_k, z_k)$ is computed as:
\begin{equation}
 d_k=\sum_{u, v} c_k(u, v), \qquad (x_k, y_k, z_k)=\sum_{u, v} m_k^{i=\{x,y,z\}}(u, v)\cdot h_k(u, v).
\vspace{\eqmargin}
\end{equation}

\Paragraph{Category-specific keypoints estimation.}
Given a category $\mathcal{T}_{c}$, our viewpoint estimation block must detect its unique $N_c$ semantic keypoints via a category-specific feature extractor $f_{\theta_{\it cat}}$ followed by a set of category-specific semantic keypoint detectors $\{f_{\theta_{\it key_k}} | k=1 \ldots N_c\}$ (lower left of~\figref{overview}). Each keypoint detector $f_{\theta_{\it key_k}}$ detects one unique category-specific semantic keypoint $k$, while the feature extractor $f_{\theta_{\it cat}}$ computes the common features required by all of them. Since our viewpoint estimation block must adapt to multiple different categories with different numbers of semantic keypoints, it cannot have a fixed number of pre-defined keypoint detectors. To flexibly change the number of keypoint detectors for each novel category, we propose a meta-Siamese architecture (lower left of~\figref{overview}), which we operate as follows. For each new category $\mathcal{T}_{c}$, we replicate a generic pre-trained keypoint detector ($f_{\theta_{\it key}}$) $N_c$ times and train each copy to detect \emph{one} unique keypoint $k$ of the new category, thus creating a specialized keypoint-detector with a unique and different number of semantic keypoints $\{f_{\theta_{\it key_k}} | k=1 \ldots N_c\}$ for each new category.

\vspace{\subsecmargin}
\subsubsection{Training}
\label{sec:metaTrain}
\vspace{\secmargin}

Our goal is to train the viewpoint estimation block to be an effective \textit{few-shot learner}. In other words, its learned feature extractor $f_{\theta_{\it cat}^*}$ and semantic keypoint detector $f_{\theta_{\it key}^*}$, after being fine-tuned with a few examples of a new category (lower left in \figref{overview}), should effectively extract features for the new category and detect each of its unique keypoints, respectively. 
To learn the optimal weights $\theta^* = \{\theta^*_{\it cat}, \theta^*_{\it key}\}$ that make our viewpoint estimation block amenable to few-shot fine-tuning without catastrophically over-fitting for a new category, we adopt the MAML meta-learning algorithm~\cite{finn2017maml}.

MAML optimizes a special meta-objective using a standard optimization algorithm, \eg SGD. 
In standard supervised learning the objective is to minimize only the \emph{training} loss for a task during each iteration of optimization. However, the meta-objective in MAML is to explicitly minimize, during each training iteration, the \emph{generalization} loss for a task \emph{after} a network has been trained with a \emph{few} of its labeled examples. Furthermore, it samples a random task from a set of many such related tasks available for training during each iteration. We describe our specific meta-traning algorithm to learn the optimal weights $\theta^* = \{\theta^*_{\it cat}, \theta^*_{\it key}\}$ for our viewpoint estimation block as follows.

During each iteration of meta-training, we sample a random task from $S^\mathrm{train}$. A task comprises of a support set ${D}^s_c$ and a query set ${D}^q_c$, each containing 10 and 3 labeled examples, respectively, of a category $\mathcal{T}_{c}$. The term ``shot'' refers to the number of examples in the support set ${D}^s_c$. For this category, containing $N_c$ semantic keypoints, we replicate our generic keypoint detector ($f_{\theta_{\it {key}}}$) $N_c$ times to construct its unique meta-Siamese keypoints detector with the parameters $\tilde\theta  \gets \left\lbrack \theta_{\it cat},\theta_{{\it key}_1},\theta_{{\it key}_2}, \ldots, \theta_{{\it key}_{N_c}} \right\rbrack$ (lower left in \figref{overview}) and initialize each $\theta_{\it key_{k}}$ with $\theta_{\it key}$. We use the category-specific keypoint detector to estimate its support set's semantic keypoints and given their ground truth values, we compute the following loss:
\begin{equation}
 \label{eq:suploss}
 \mathcal{L}_{\mathcal{T}_c}^s=\lambda_\mathrm{2D}\mathcal{L}_\mathrm{2D} + \lambda_\mathrm{3D}\mathcal{L}_\mathrm{3D} + \lambda_\mathrm{d}\mathcal{L}_\mathrm{d},
\vspace{\eqmargin}
\end{equation}
where $\mathcal{L}_\mathrm{2D}$, $\mathcal{L}_\mathrm{3D}$, 
and $\mathcal{L}_\mathrm{d}$ are the average $L_2$ regression losses for correctly estimating the semantic keypoints' 2D and 3D positions, and depth estimates, respectively.
The $\lambda$ parameters control the relative importance of each loss term. We compute the gradient of this loss $\mathcal{L}_{\mathcal{T}_c}^s$ w.r.t.\ to the network's parameters $\tilde\theta$ and use a single step of SGD to update $\tilde\theta$ to $\tilde\theta'$ with a learning rate of $\alpha$:
\begin{equation}
\tilde\theta' \gets \tilde\theta-\alpha\nabla_{\tilde\theta}{\mathcal{L}_{\mathcal{T}_c}^s}.
\vspace{-0.5mm}
\end{equation} 

Next, with the updated model parameters $\tilde\theta'$, we compute the loss $\mathcal{L}_{\mathcal{T}_c}^q$ for the query set $D^q_c$ of this category (lower right in \figref{overview}). To compute the query loss, in addition to the loss terms described in~\eqnref{suploss}, we also use a weighted concentration loss term:
\vspace{\subsecmargin}
\begin{equation}
 \mathcal{L}_{con} = \frac{1}{N_c}\sum_{k=1}^{N_c}\sum_{u,v}h_k(u,v)\|\left[u_k,v_k\right]^\top-\left[u,v\right]^\top\|_2,
\vspace{\eqmargin}
\end{equation}

\noindent which forces the distribution of a 2D keypoint's heatmap $h_k(u,v)$ to be peaky around the predicted position $(u_k, v_k)$. We find that this concentration loss term helps to improve the accuracy of 2D keypoint detection. Our final query loss is:
\begin{equation}
 \label{eq:queryloss}
 \mathcal{L}_{\mathcal{T}_c}^q=\lambda_\mathrm{2D}\mathcal{L}_\mathrm{2D} + \lambda_\mathrm{3D}\mathcal{L}_\mathrm{3D} + \lambda_\mathrm{d}\mathcal{L}_\mathrm{d} + \lambda_\mathrm{con}\mathcal{L}_\mathrm{con}.
\vspace{\eqmargin}
\end{equation}
\noindent The generalization  loss of our network $\mathcal{L}_{\mathcal{T}_c}^q$, after it has been trained with just a few examples of a specific category, serves as the final meta-objective that is minimized in each iteration of meta-training and we optimize the network's initial parameters $\theta$ w.r.t.\ its query loss $\mathcal{L}_{\mathcal{T}_c}^q$ using:

\begin{equation}
\label{eq:catUpdate}
\theta_{\it cat} \gets \theta_{\it cat}-\beta\nabla_{\theta_{\it cat}} 
    \mathcal{L}_{\mathcal{T}_{c}}^q \left(f_{\tilde\theta'}\right),
\vspace{\eqmargin}
\end{equation}

\begin{equation}
\label{eg:keypointUpdate}
\theta_{\it key} \gets \theta_{\it key}-\beta 
                \frac{1}{N_{c}}
                \sum\limits_{k = 1..N_c} 
                \left\lbrack
                \nabla_{\theta_{{\it key}_k}} 
                \mathcal{L}_{\mathcal{T}_{c}}^q\hspace{-1.5mm}
                \left(f_{\tilde\theta'}
                \right)
                \right\rbrack.
\vspace{-0.5mm}
\end{equation}

We repeat the meta-training iterations until our viewpoint estimation block converges to $f_{\theta^{*}}$, as presented in Algorithm~\ref{alg:metapose}. Notice that in Eq.~(\ref{eg:keypointUpdate}) we compute the optimal weights for the generic keypoint detector $\theta_{\it key}$ by averaging the gradients of all the duplicated keypoint detectors $\theta_{\it key_k}$. We find that this novel design feature of our network along with its shared category-level feature extractor with parameters $\theta_{\it cat}$ help to improve accuracy. They enable efficient use of \emph{all} the available keypoints to learn the optimal values for  $\theta_{\it cat}$ and $\theta_{\it key}$ during meta-training, which is especially important when training data is scarce.

\vspace{-1.5mm}
\begin{algorithm}
  \caption{MetaView Meta-training}
  \label{alg:metapose}
  \begin{algorithmic}[1]
    \State \textbf{Require:} a set of tasks $S^\mathrm{train}$ \label{alg:firstline}
    \State randomly initialize $\theta_{\it key}$ and $\theta_{\it cat}$
    \While{training}
    \State sample one task $\mathcal{T}_c \sim S^\mathrm{train}$
        \State $\triangleright$  meta-Siamese keypoint detectors
        \State $\theta_{{\it key}_1},\theta_{{\it key}_2}, \ldots, \theta_{{\it key}_{N_c}} \gets \theta_{\it key}$ \label{alg:copy}
        \State $\triangleright$  viewpoint estimator
        \State    $\tilde\theta  \gets \left\lbrack \theta_{\it cat},\theta_{{\it key}_1},\theta_{{\it key}_2}, \ldots, \theta_{{\it key}_{N_c}} \right\rbrack$ \label{alg:merge}
        \State $\triangleright$  update viewpoint estimator using support set
        \State    $\tilde\theta' \gets \tilde\theta-\alpha\nabla_{\tilde\theta}{\mathcal{L}_{\mathcal{T}_c}^s}\hspace{-1.5mm}
        \left(f_{\tilde\theta}\right)
         $\label{alg:update1}

        \State $\triangleright$ meta learning optimization using query set
        \State
            $\theta_{\it cat} \gets \theta_{\it cat}-\beta\nabla_{\theta_{\it cat}} 
                \mathcal{L}_{\mathcal{T}_{c}}^q\hspace{-1.5mm}
                \left(f_{\tilde\theta'}\right)        
            $\label{alg:update2}
        \State
            $\theta_{\it key} \gets \theta_{\it key}-\beta 
                \frac{1}{N_{c}}
                \sum\limits_{k = 1..N_c} 
                \left\lbrack
                \nabla_{\theta_{{\it key}_k}} 
                \mathcal{L}_{\mathcal{T}_{c}}^q\hspace{-1.5mm}
                \left(f_{\tilde\theta'}
                \right)
                \right\rbrack
            $\label{alg:update3}
    \EndWhile
  \end{algorithmic}
\end{algorithm}
\vspace{-5mm}

\vspace{\secmargin}
\subsubsection{Inference}
\vspace{\secmargin}

 We evaluate the performance of how well our viewpoint estimation block $f_{\theta^*}$, which is learned via meta-learning performs at the task of adapting to unseen categories. Similar to meta-training, we sample a category from $S^\mathrm{test}$ with the same shot size as used for training. We construct its unique viewpoint estimation network $f_{\tilde\theta^*}$ and fine-tune it with a few of its examples by minimizing the loss in Eq.~(\ref{eq:suploss}). This results in a optimal few-shot trained network $f_{\tilde{\theta}^{*'}}$ for this category. We then evaluate the generalization performance of $f_{\tilde{\theta}^{*'}}$ on all testing images of that category. We repeat this procedure for all categories in $S^\mathrm{test}$ and for multiple randomly selected few-shot training samples per category, and average across all of them.
\vspace{-0.3cm}

\vspace{-1mm}
\vspace{\secmargin}
\section{Results}
\vspace{\secmargin}

\Paragraph{Implementation details.} 
We provide detailed descriptions of our CNN architectures, and their training procedures in the supplementary material, to limit the number of pages.

\Paragraph{Experiments.}\label{paragraph:settings}
We evaluate our method for two different experimental settings.
First, we follow the \textit{intra}-dataset experiment of~\cite{zhou2018starmap} and split the categories in ObjectNet3D~\cite{xiang2016objectnet3d} into $76$ and $17$ for training and testing, respectively.
Secondly, we conduct an \textit{inter}-dataset experiment.
From ObjectNet3D, we exclude the 12 categories that are also present in Pascal3D+~\cite{xiang_pascal3d}.
We then use the remaining $88$ categories in ObjectNet3D for training and test on Pascal3D+.
Complying with~\cite{tulsiani2015viewpoints}, we discard the images with occluded or truncated objects from the test set in both experiments.
We use two metrics for evaluation: 1) \textit{Acc30}, which is the percentage of views with a rotational error less than $30\degree$ and 2) \textit{MedErr}, which is the median rotational error across a dataset, measured in degrees.
We compute the rotational error as $E_R=\frac{\|\log(R_{gt}^\top R)\|_F}{\sqrt{2}}$, where $\|\cdot\|_F$ is the Frobenius norm, and $R_{gt}$ and $R$ are the ground truth and predicted rotation matrices, respectively.

\begin{table}[t]\tiny\addtolength{\tabcolsep}{-1pt}
    \centering
    \caption{\textbf{Intra-dataset experiment}.
    We report \textit{Acc30}($\uparrow$)/\textit{MedErr}($\downarrow$). 
    All models are trained and evaluated on 76 and 17 categories from ObjectNet3D, respectively.
    The ``zero'' methods don't use images of unknown categories for training and all others involve few-shot learning.
    }
    \label{tab:intra}
    \begin{tabular}{l cccccccccc} 
        \toprule
	    Method & bed & bookshelf & calculator & cellphone & computer & f$\_$cabinet & guitar & iron & knife & microwave \\
        \midrule
        StarMap (zero) & 0.37 / 45.1 & 0.69 / 18.5 & 0.19 / 61.8 & 0.51 / 29.8 & 0.74 / 15.6 & 0.78 / 14.1 & 0.64 / 20.4 & 0.02 / 142 & 0.08 / 136 & 0.89 / 12.2 \\
        StarMap* (zero) & 0.31 / 45.0 & 0.63 / 22.2 & 0.27 / 52.2 & 0.51 / 29.8 & 0.64 / 24.2 & 0.78 / 15.8 & 0.52 / 28.0 & 0.00 / 134 & 0.06 / 124 & 0.82 / 16.9 \\
        Baseline (zero)& 0.26 / 49.1 & 0.57 / 25.0 & 0.78 / 53.3 & 0.38 / 45.5 & 0.66 / 20.3 & 0.73 / 18.7 & 0.39 / 44.6 & 0.06 / 135 & 0.08 / 127 & 0.82 / 16.8  \\
        \midrule
        StarMap* + fine-tune & 0.32 / 47.2 & 0.61 / 21.0 & 0.26 / 50.6 & 0.56 / 26.8 & 0.59 / 24.4 & \textbf{0.76} / 17.1 & 0.54 / 27.9 & 0.00 / 128 & 0.05 / 120 & 0.82 / 19.0 \\
        Baseline + fine-tune& 0.28 / 43.7 & 0.67 / 22.0 & 0.77 / 18.4 & 0.45 / 34.6 & 0.67 / 22.7 & 0.67 / 21.5 & 0.27 / 52.1 & 0.02 / 127 & 0.06 / 108 & \textbf{0.85} / \textbf{16.6} \\
        StarMap* + MAML & 0.32 / 42.2 & \textbf{0.76} / \textbf{15.7} & 0.58 / 26.8 & \textbf{0.59} / \textbf{22.2} & 0.69 / \textbf{19.2} & \textbf{0.76} / \textbf{15.5} & 0.59 / \textbf{21.5} & 0.00 / 136 & \textbf{0.08} / 117 & 0.82 / 17.3 \\
        Ours & \textbf{0.36} / \textbf{37.5} & \textbf{0.76} / 17.2 & \textbf{0.92} / \textbf{12.3} & 0.58 / 25.1 & \textbf{0.70} / 22.2 & 0.66 / 22.9 & \textbf{0.63} / 24.0 & \textbf{0.20} / \textbf{76.9} & 0.05 / \textbf{97.9} & 0.77 / 17.9 \\
        \midrule
        \midrule
        Method & pot & rifle & slipper & stove & toilet & tub & wheelchair &  \multicolumn{3}{c}{\cellcolor{lightgray}TOTAL} \\
        \midrule
        StarMap (zero) & 0.50 / 30.0 & 0.00 / 104 & 0.11 / 146 & 0.82 / 12.0 & 0.43 / 35.8 & 0.49 / 31.8 & 0.14 / 93.8 &  \multicolumn{3}{c}{0.44 / 39.3} \\
        StarMap* (zero) & 0.51 / 29.2 & 0.02 / 97.4 & 0.10 / 130 & 0.81 / 13.9 & 0.44 / 34.4 & 0.37 / 37.0 & 0.17 / 74.4 &  \multicolumn{3}{c}{0.43 / 39.4} \\
        Baseline (zero) & 0.46 / 38.8 & 0.00 / 98.6 & 0.09 / 123 & 0.82 / 14.8 & 0.32 / 39.5 & 0.29 / 50.4 & 0.14 / 71.6 & \multicolumn{3}{c}{0.38 / 44.6} \\
        \midrule
        StarMap* + fine-tune & \textbf{0.51} / 29.9 & 0.02 / 100 & 0.08 / 128 & 0.80 / 16.1 & 0.38 / 36.8 & 0.35 / 39.8 & 0.18 / 80.4 & \multicolumn{3}{c}{\cellcolor{lightgray}0.41 $\pm$ 0.00 / 41.0 $\pm$ 0.22} \\
        Baseline + fine-tune & 0.38 / 39.1 & 0.01 / 107 & 0.03 / 123 & 0.72 / 21.6 & 0.31 / 39.9 & 0.28 / 48.5 & 0.15 / 70.8 & \multicolumn{3}{c}{\cellcolor{lightgray}0.40 $\pm$ 0.02 / 39.1 $\pm$ 1.79} \\
        StarMap* + MAML & \textbf{0.51} / \textbf{28.2} & 0.01 / 100 & \textbf{0.15} / 128 & \textbf{0.83} / \textbf{15.6} & 0.39 / 35.5& \textbf{0.41} / \textbf{38.5} & 0.24 / 71.5 &  \multicolumn{3}{c}{\cellcolor{lightgray}0.46 $\pm$ 0.01 / 33.9 $\pm$ 0.16} \\
        Ours & 0.49 / 31.6 & \textbf{0.21} / \textbf{80.9} & 0.07 / \textbf{115} & 0.74 / 21.7 & \textbf{0.50} / \textbf{32.0} & 0.29 / 46.5 & \textbf{0.27} / \textbf{55.8} &  \multicolumn{3}{c}{\cellcolor{lightgray}\textcolor{red}{\textbf{0.48}} $\pm$ 0.01 / \textcolor{red}{\textbf{31.5}} $\pm$ 0.72} \\
        \bottomrule
    \end{tabular}
    \vspace{\tabmargin}
\end{table}
\begin{table}[t]\semitiny
    \centering
    \caption{\textbf{Inter-dataset experiment}.
    We report \textit{Acc30}($\uparrow$)/\textit{MedErr}($\downarrow$).
    All models are trained on ObjectNet3D and evaluated on Pascal3D+.
    The ``zero'' methods don't use images of unknown categories for training and all others involve few-shot learning.
    }
    \label{tab:inter}
    \begin{tabular}{l ccccccc} 
        \toprule
	    Method & aero & bike & boat & bottle & bus & car & chair \\
	    \midrule
        StarMap (zero) & 0.04 / 97.7 & 0.10 / 90.42 & 0.14 / 78.42 & 0.81 / 16.7 & 0.54 / 29.4 & 0.25 / 67.8 & 0.19 / 97.3 \\
        StarMap* (zero) & 0.02 / 112 & 0.02 / 102 & 0.06 / 110 & 0.44 / 34.3 & 0.48 / 32.7 & 0.18 / 87.0 & 0.29 / 70.0 \\
        Baseline (zero) & 0.03 / 114 & 0.06 / 101 & 0.10 / 95 & 0.41 / 36.6 & 0.36 / 42.0 & 0.14 / 93.7 & 0.26 / 71.5 \\
        \midrule
        StarMap* + fine-tune & 0.03 / 102 & 0.05 / 98.8 & 0.07 / 98.9 & 0.48 / 31.9 & 0.46 / 33.0 & 0.18 / 80.8 & \textbf{0.22} / \textbf{74.6} \\
        Baseline + fine-tune & 0.02 / 113 & 0.04 / 112 & \textbf{0.11} / 93.4 & 0.39 / 37.1 & 0.35 / 39.9 & 0.11 / 99.0 & 0.21 / 75.0 \\
        StarMap* + MAML & 0.03 / \textbf{99.2} & \textbf{0.08} / \textbf{88.4} & \textbf{0.11} / \textbf{92.2} & 0.55 / 28.0 & 0.49 / 31.0 & 0.21 / 81.4 & 0.21 / 80.2  \\
        Ours & \textbf{0.12} / 104 & \textbf{0.08} / 91.3 & 0.09 / 108 & \textbf{0.71} / \textbf{24.0} & \textbf{0.64} / \textbf{22.8} & \textbf{0.22} / \textbf{73.3} & 0.20 / 89.1 \\
        \midrule
        \midrule
        Method & table & mbike & sofa & train & tv & \multicolumn{2}{c}{\cellcolor{lightgray}TOTAL} \\
	    \midrule
        StarMap (zero) & 0.62 / 23.3 & 0.15 / 70.0 & 0.23 / 49.0 & 0.63 / 25.7 & 0.46 / 31.3 & \multicolumn{2}{c}{0.32 / 50.1} \\
        StarMap* (zero) & 0.43 / 31.7 & 0.09 / 86.7 & 0.26 / 42.5 & 0.30 / 46.8 & 0.59 / 24.7 & \multicolumn{2}{c}{0.25 / 71.2} \\
        Baseline (zero) & 0.38 / 39.0 & 0.11 / 82.3 & 0.39 / 57.5 & 0.29 / 50.0 & 0.63 / 24.3 & \multicolumn{2}{c}{0.24 / 70.0} \\
        \midrule
        StarMap* + fine-tune & \textbf{0.46} / \textbf{31.4} & 0.09 / 91.6 & 0.32 / 44.7 & 0.36 / 41.7 & 0.52 / 29.1 & \multicolumn{2}{c}{\cellcolor{lightgray}0.25 $\pm$ 0.01 / 64.7 $\pm$ 1.07} \\
        Baseline + fine-tune & 0.41 / 35.1 & 0.09 / 79.1 & 0.32 / 58.1 & 0.29 / 51.3 & 0.59 / 29.9 & \multicolumn{2}{c}{\cellcolor{lightgray}0.22 $\pm$ 0.02 / 69.2 $\pm$ 1.48} \\
        StarMap* + MAML & 0.29 / 36.8 & 0.11 / 83.5 & \textbf{0.44} / \textbf{42.9} & 0.42 / 33.9 & \textbf{0.64} / \textbf{25.3} & \multicolumn{2}{c}{\cellcolor{lightgray}0.28 $\pm$ 0.00 / 60.5 $\pm$ 0.10} \\
        Ours & 0.39 / 36.0 & \textbf{0.14} / \textbf{74.7} & 0.29 / 46.2 & \textbf{0.61} / \textbf{23.8} & 0.58 / 26.3 & \multicolumn{2}{c}{\cellcolor{lightgray}\textcolor{red}{\textbf{0.33}} $\pm$ 0.02 / \textcolor{red}{\textbf{51.3}} $\pm$ 4.28} \\
        \bottomrule
    \end{tabular}
    \vspace{\tabmargin}
\end{table}

\begin{table}[t]\tiny\addtolength{\tabcolsep}{-1pt}
    \centering
    \caption{\textbf{Ablation study}.
    The table shows the individual contributions of our meta-Siamese design (MS), the concentration loss ($L_{con}$), and general keypoints heatmap (KP) on the performance of MetaView in the intra-dataset experiment.
    We report \textit{Acc30}($\uparrow$)/\textit{MedErr}($\downarrow$).
    }
    \label{tab:ablation}
    \begin{tabular}{@{}l cccccccccc@{}} 
        \toprule
	    Method & bed & bookshelf & calculator & cellphone & computer & f$\_$cabinet & guitar & iron & knife & microwave \\
	    \midrule
	    Ours & 0.28 / 42.3 & 0.68 / 23.1 & 0.87 / 15.3 & 0.47 / 32.1 & 0.63 / 24.9 & \textbf{0.71} / \textbf{22.1} & 0.03 / 100 & 0.15 / 76.0 & 0.01 / 121 & 0.69 / 23.2 \\
        Ours (MS) & 0.27 / 42.4 & 0.77 / 22.2 & 0.74 / 24.0 & 0.54 / 28.3 & 0.64 / 24.9 & 0.63 / 25.3 & 0.61 / 25.3 & 0.13 / \textbf{76.9} & 0.05 / 103 & 0.65 / 26.2 \\
        Ours (MS, $\mathcal{L}_\mathrm{con}$) & 0.31 / 41.3 & \textbf{0.79} / 19.0 & 0.84 / 17.4 & 0.53 / 28.0 & 0.62 / 25.9 & 0.66 / 23.6 & 0.35 / 35.8 & 0.16 / 86.5 & 0.05 / 101 & \textbf{0.81} / \textbf{17.7} \\
        Ours (MS, $\mathcal{L}_\mathrm{con}$, KP) & \textbf{0.36} / \textbf{37.5} & 0.76 / \textbf{17.2} & \textbf{0.92} / \textbf{12.3} & \textbf{0.58} / 25.1 & \textbf{0.70} / \textbf{22.2} & 0.66 / 22.9 & \textbf{0.63} / \textbf{24.0} & \textbf{0.20} / \textbf{76.9} & 0.05 / \textbf{97.9} & 0.77 / 17.9 \\
        \midrule
        \midrule
        Method & pot & rifle & slipper & stove & toilet & tub & wheelchair & \multicolumn{3}{c}{\cellcolor{lightgray}TOTAL} \\
        \midrule
        Ours & 0.46 / 32.1 & 0.04 / 119 & 0.02 / 125 & \textbf{0.81} / \textbf{19.5} & 0.15 / 51.2 & 0.26 / 45.9 & 0.02 / 109 & \multicolumn{3}{c}{\cellcolor{lightgray}0.35 $\pm$ 0.01 / 42.5 $\pm$ 1.15} \\
        Ours (MS) & 0.34 / 37.4 & 0.18 / \textbf{78.8} & 0.05 / \textbf{111} & 0.71 / 21.5 & 0.37 / 35.8 & 0.24 / 44.8 & 0.10 / 76.1 &  \multicolumn{3}{c}{\cellcolor{lightgray}0.41 $\pm$ 0.01 / 36.0 $\pm$ 0.78} \\
        Ours (MS, $\mathcal{L}_\mathrm{con}$) & \textbf{0.49} / \textbf{31.2} & 0.16 / 90.5 & 0.05 / \textbf{111} & 0.75 / 21.7 & 0.41 / 34.4 & 0.31 / 42.4 & 0.22 / 60.8 & \multicolumn{3}{c}{\cellcolor{lightgray}0.45 $\pm$ 0.01 / 33.6 $\pm$ 0.94}\\
        Ours (MS, $\mathcal{L}_\mathrm{con}$, KP) & \textbf{0.49} / 31.6 & \textbf{0.21} / 80.9 & \textbf{0.07} / 115 & 0.74 / 21.7 & \textbf{0.50} / 32.0 & 0.29 / 46.5 & \textbf{0.27} / \textbf{55.8} &  \multicolumn{3}{c}{\cellcolor{lightgray}\textcolor{red}{\textbf{0.48}} $\pm$ 0.01 / \textcolor{red}{\textbf{31.5}} $\pm$ 0.72} \\
        \bottomrule
    \end{tabular}
    \vspace{\tabmargin}
\end{table}

\Paragraph{Comparisons.}
We compare several viewpoint estimation networks to ours. These include:
\begin{compactitem}

\item \tb{StarMap}: The original StarMap method~\cite{zhou2018starmap}. It contains two stages of an Hourglass network~\cite{Newell2016eccv:posehuman} as the backbone and computes a multi-peak heatmap of general visible keypoints, and their depth and canonical 3D points. 

\item \tb{StarMap*}: Our re-implementation of StarMap~\cite{zhou2018starmap} with one stage of ResNet-18~\cite{he2016deep} as the backbone for a fair comparison to ours.

\item \tb{StarMap* + MAML}: The StarMap* network trained with MAML for few-shot viewpoint estimation.

\item \tb{Baseline}: The ResNet-18 network trained to detect a fixed number~(8) of semantic keypoints for all categories via standard supervised learning.

\end{compactitem}
For methods that involve few-shot fine-tuning on unknown categories (\textit{i.e.}, StarMap* or Baseline with fine-tuning, StarMap + MAML, and Ours), we use a shot size of $10$.
We repeat each experiment ten times with random initial seeds and report their average performance.
%
Note that we also attempted to train viewpoint estimation networks that estimate angular values directly (\eg~\cite{xiang2018rss:posecnn}); or those that detect projections of 3D bounding boxes (\eg~\cite{grabner20183d}) with MAML, but they either failed to converge to performed very poorly. So, we do not report results for them. The results of the intra-dataset and inter-dataset experiments are presented in ~\tabref{intra} and ~\tabref{inter}, respectively.

\Paragraph{Zero-shot performance.} 
For both experiments, methods trained using standard supervised learning solely on the training categories (\textit{i.e.}, StarMap, StarMap* and Baseline denoted by ``zero'') are limited in their ability to generalize to unknown categories.
For the original StarMap method~\cite{zhou2018starmap} in the intra-dataset experiment (\tabref{intra}), the overall \textit{Acc30} and \textit{MedErr} worsen from $63\%$ and $17^\circ$, respectively, when the test categories are known to the system to $44\%$ and $39.3^\circ$, respectively, when they are unknown. 
This indicates that the existing state-of-the-art viewpoint estimation networks require information that is unique to each category to infer its viewpoint. Since the original StarMap~\cite{zhou2018starmap} uses a larger backbone network than ResNet-18~\cite{he2016deep} it performs better than our implementation (StarMap*) of it.

\begin{table*}\tiny\addtolength{\tabcolsep}{-1pt}
    \centering
    \caption{\textbf{Shot size}.
    We report \textit{Acc30}($\uparrow$)/\textit{MedErr}($\downarrow$).
    The table shows the effect of varying the number of support images (``shot size'') during meta-training and testing in the intra-dataset experiments with ObjectNet3D.
    }
    \label{tab:shot}
    \begin{tabular}{l cccccccccc} 
        \toprule
	    Method & bed & bookshelf & calculator & cellphone & computer & f$\_$cabinet & guitar & iron & knife & microwave \\
	    \midrule
	    Ours (1 shot) & 0.24 / 45.7 & 0.16 / 70.8 & 0.26 / 56.7 & 0.19 / 57.3 & 0.41 / 32.6 & 0.48 / 31.4 & 0.06 / 76.8 & 0.02 / 125 & 0.01 / 120 & 0.18 / 48.4 \\
        Ours (5 shots) & 0.31 / 39.9 & 0.50 / 29.6 & 0.67 / 25.1 & 0.34 / 48.6 & 0.67 / 23.7 & 0.66 / 24.0 & 0.34 / 40.4 & 0.09 / 91.7 & 0.04 / 110 & 0.81 / 16.7 \\
        Ours (10 shots) & \textbf{0.36} / \textbf{37.5} & 0.76 / \textbf{17.2} & \textbf{0.92} / \textbf{12.3} & \textbf{0.58} / 25.1 & \textbf{0.70} / \textbf{22.2} & 0.66 / 22.9 & \textbf{0.63} / \textbf{24.0} & \textbf{0.20} / \textbf{76.9} & 0.05 / \textbf{97.9} & 0.77 / 17.9 \\
        \midrule
        \midrule
        Method & pot & rifle & slipper & stove & toilet & tub & wheelchair & \multicolumn{3}{c}{\cellcolor{lightgray}TOTAL} \\
        \midrule
        Ours (1 shot) & 0.39 / 36.8 & 0.00 / 102 & 0.05 / 121 & 0.36 / 35.8 & 0.33 / 39.1 & 0.11 / 75.1 & 0.12 / 81.5 & \multicolumn{3}{c}{\cellcolor{lightgray}0.21 $\pm$ 0.05 / 55.2 $\pm$ 6.82}\\
        Ours (5 shots) & 0.51 / 29.4 & 0.05 / 107 & 0.04 / 110 & 0.74 / 21.1 & 0.38 / 35.7 & 0.27 / 46.7 & 0.23 / 61.3 & \multicolumn{3}{c}{\cellcolor{lightgray}0.41 $\pm$ 0.03 / 36.2 $\pm$ 1.58} \\
        Ours (10 shots) & \textbf{0.49} / 31.6 & \textbf{0.21} / 80.9 & \textbf{0.07} / 115 & 0.74 / 21.7 & \textbf{0.50} / 32.0 & 0.29 / 46.5 & \textbf{0.27} / \textbf{55.8} &  \multicolumn{3}{c}{\cellcolor{lightgray}\textcolor{red}{\textbf{0.48}} $\pm$ 0.01 / \textcolor{red}{\textbf{31.5}} $\pm$ 0.72} \\
        \bottomrule
    \end{tabular}
    \vspace{\tabmargin}
\end{table*}

\Paragraph{Few-shot performance.} 
Among the methods that involve few-shot fine-tuning for unknown categories, methods that are trained via meta-learning (StarMap + MAML and our MetaView) perform significantly better than the methods that are not (StarMap* or Baseline with fine-tuning) in both the intra- and inter-dataset experiments.
%
%
These results are the first demonstration of the effectiveness of meta-learning at the task of category-level few-shot viewpoint learning.
Furthermore, in both experiments, our MetaView framework results in the best overall performance of all the zero- and few-shot learning methods. 
It outperforms StarMap* + MAML, which shows the effectiveness of our novel design components that differentiate it from merely training StarMap* with MAML.
They include our network's ability to (a) detect the 2D locations and depth values of all 3D canonical points and not just the visible ones; (b) share information during meta-learning via the meta-Siamese design; and (c) flexibly construct networks with a different number of keypoints for each category. 
Lastly, observe that even with a smaller backbone network, our method performs better than the current best performance for the task of viewpoint estimation of unknown categories, \textit{i.e.} of StarMap~\cite{zhou2018starmap} ``zero'' and thus helps to improve performance on unknown categories with very little additional labeling effort.

\begin{figure*}
	\centering
    \includegraphics[width=0.9\linewidth]{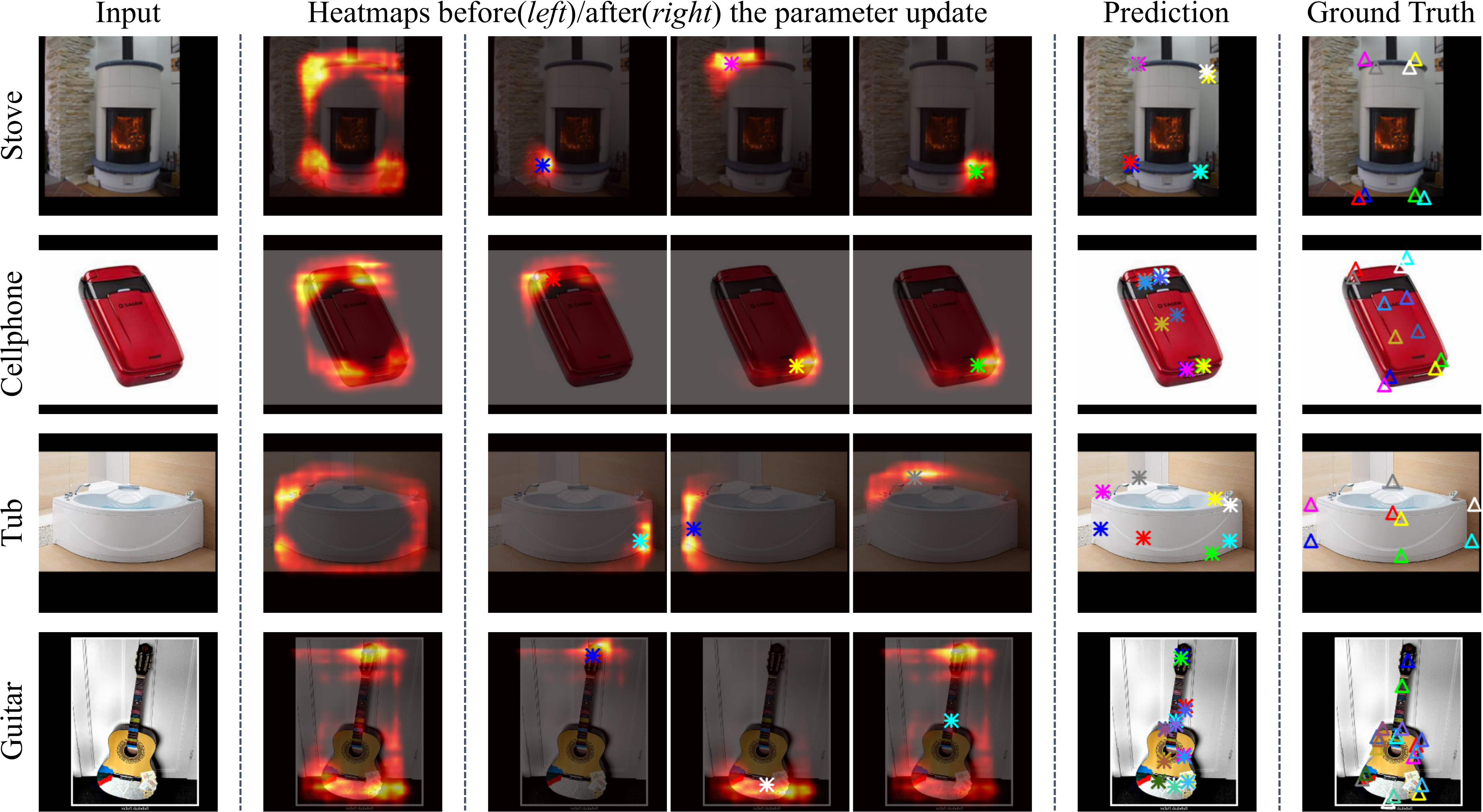}
    \vspace{-2mm}
    \caption{\textbf{Qualitative results of the intra-dataset experiment.} We show the keypoint detection results of MetaView on unknown categories, before and after few-shot fine-tuning. The images from left to right are: the input image, the 2D keypoint heatmap before fine-tuning with Eq.~\eqref{eq:suploss}, three example heatmaps for specific keypoints after fine-tuning, all the predicted keypoints, and their ground truth values.}
    \label{figure:qualitative_object3d}
    \vspace{\figmargin}
 \end{figure*}

The effectiveness of MetaView is also evident from~\figref{qualitative_object3d}, which shows examples of the 2D keypoint heatmaps $h_k(u, v)$ (described in~\secref{3_2}) produced by it before and after few-shot fine-tuning with examples of new categories. The keypoint detector, prior to few-shot fine-tuning, is not specific to any keypoint and generates heatmaps that tend to have high responses on corners, edges or regions of the foreground object. After fine-tuning, however, it successfully learns to detect keypoints of various new categories and produces heatmaps with more concentrated peaks.

\Paragraph{Ablation study.} To validate the effectiveness of our various novel design components including our meta-Siamese design, concentration loss term ($L_{con}$), and of using the general keypoints' multi-peak map as input, we show the results of an ablation study for the inter-dataset experiment in~\tabref{ablation}.
While each component individually contributes to the overall performance, the concentration loss and the meta-Siamese design contribute the most.
%

\Paragraph{Shot size.}
We vary the number of support images (\textit{i.e.}, shot size to $1$, $5$ and $10$) for each new category during meta-training and -testing. The results of this experiment for the intra-dataset setting are presented in Table~\ref{tab:shot}. We observe that as more training images per category are available for training, the accuracy of our MetaView approach scales up correspondingly. 
\vspace{\secmargin}
\vspace{-1.5mm}
\section{Conclusion}
\vspace{\secmargin}
\vspace{-1mm}
To improve performance on unknown categories, we introduce the problem of category-level few-shot viewpoint estimation. We propose the novel  MetaView framework that successfully adapts to unknown categories with few labeled examples and helps to improve performance on them with little additional annotation effort. Our meta-Siamese keypoint detector is general and can be explored in the future for other few-shot tasks requiring keypoints detection.

\bibliography{egbib}

\clearpage
\begin{appendices}
\section{Implementation Details}

\Paragraph{Network architecture.}
The proposed MetaView framework contains a feature extraction block and a viewpoint estimation block.
The feature extraction block consists of an image feature extractor and a general keypoints detector, while the viewpoint estimation block comprises of a category-specific feature extractor $f_{\theta_{cat}}$ and a keypoint detector $f_{\theta_{key}}$.
We present the architectural details of each module in~\tabref{arch}.

\Paragraph{Training details.}
We implement our model using PyTorch~\cite{paszke2017pytorch}.
The size of an input image is $256\times256$, and the number of images in the support set $N_s$ and query set $N_q$ are $10$ and $3$, respectively.
For data augmentation, we randomly apply mirroring, translations, and rotations to images in both the support and query sets, where the interval of random rotations is set to $\left[-60\degree, 60\degree\right]$.
The learning rate for the single stochastic gradient descent optimization step (Alg.~\ref{alg:metapose}, line 10) is $0.01$.
For training, we use the Adam~\cite{kinga2015method} optimizer with an initial learning rate of 5e-4 (Alg.~\ref{alg:metapose}, lines 12 and 13).
The model is trained for $120$ epochs, and the learning rate is decayed by a factor of $0.5$ after $80$ and $110$ epochs.
Due to limitations imposed by the size of our GPU memory, we set the size of the mini-batch during meta-training to $1$ meta-task (comprising of $10$ support and $3$ query images of a single category) per iteration.
For the objective function:
\begin{equation}
\label{eq:queryloss}
\mathcal{L}^q_c = \lambda_\mathrm{2D}\mathcal{L}_\mathrm{2D} + \lambda_\mathrm{3D}\mathcal{L}_\mathrm{3D} + \lambda_\mathrm{d}\mathcal{L}_\mathrm{d} + \lambda_\mathrm{con}\mathcal{L}_\mathrm{con},
\end{equation}
we set the hyper-parameters $\lambda_\mathrm{2D}$, $\lambda_\mathrm{3D}$, $\lambda_\mathrm{d}$ and $\lambda_\mathrm{con}$ to $50$, $1$, $0.2$ and $0.5$, respectively, which we determined empirically.
In practice, we split the meta-training into two stages.
In the first stage, we train the model with the 2D keypoint position prediction loss $L_\mathrm{2D}$ and the concentration $L_\mathrm{con}$ loss only.
In the second stage, we optimize the full objective function $\mathcal{L}^q_c$ described in Equation~\eqnref{queryloss}.
We observe that this two-stage training procedure performs slightly better.

\Paragraph{Data preparation.}
We use the ObjectNet3D~\cite{xiang2016objectnet3d} and Pascal3D+~\cite{xiang_pascal3d} datasets for evaluation.
Firstly, we use the annotated ground truth 2D bounding boxes around the objects to create their cropped versions.
The ground truth 3D keypoints $(x,y,z)$ corresponding to the semantic keypoints are defined by the locations of annotated 3D anchor points on the approximate 3D CAD model that are manually aligned to the objects.
We obtain the ground truth 2D keypoint positions and depth values $(u,v,d)$ by transforming the 3D features $(x,y,z)$ from an object-centered coordinate system to the camera's viewing plane. In order to do so, we compute the projections of the 3D keypoints on to the camera's plane after rotating them with the provided ground truth rotation matrix $R_{gt}$ and applying the camera's intrinsic parameters.

\section{Additional Experiments}

\begin{figure}
	\centering
    \includegraphics[width=0.95\linewidth]{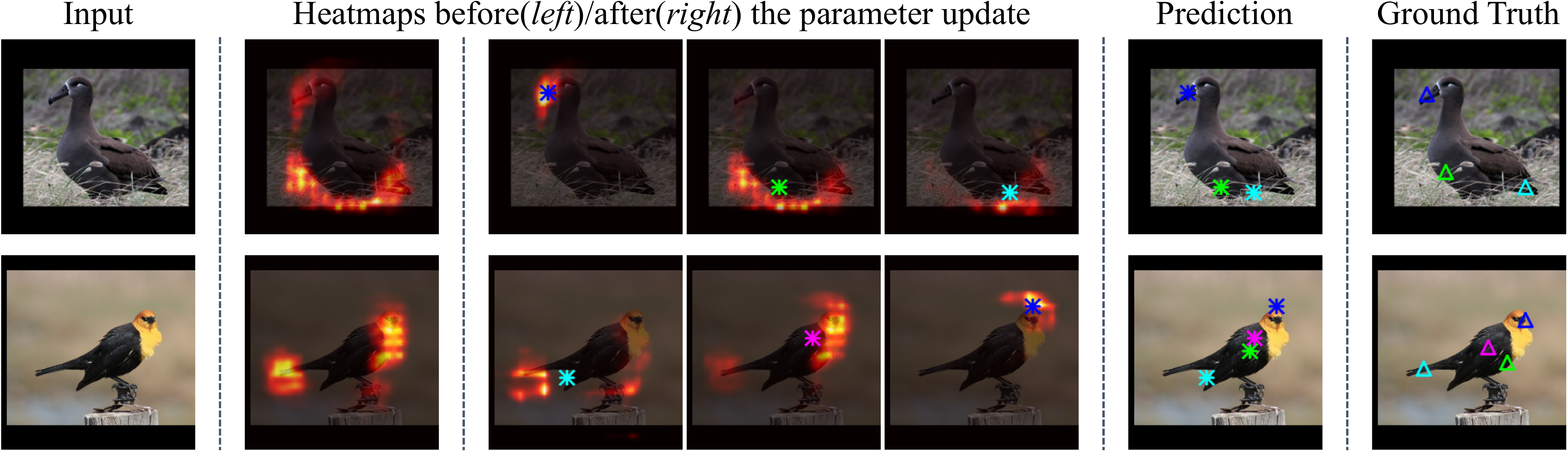}
    \caption{\textbf{Qualitative results of semantic keypoint detection on the CUB-200-2011 dataset.} We show additional results of learning to detect keypoints with MetaView. We train MetaView with $76$ categories of \textit{rigid} objects from the ObjecteNet3D~\cite{xiang2016objectnet3d} dataset, and evaluate on different categories of \textit{non-rigid} birds from the CUB-200-2011~\cite{WahCUB_200_2011} dataset. The images from left to right are: the input image, the 2D keypoint heatmap before the network's parameters are updated for a particular category, three example heatmaps for specific keypoints after the parameter update, the predicted keypoints, and the ground truth keypoints.}
    \label{figure:qualitative_cub}
 \end{figure}

\Paragraph{Non-rigid objects.}
As a preliminary investigation, we show the potential of the proposed method on detecting semantic keypoints of \textit{non-rigid} objects.
We train our model on $76$ categories of \textit{rigid} objects on the ObjectNet3D~\cite{xiang2016objectnet3d} dataset, and evaluate its performance on different categories of birds from the CUB-200-2011~\cite{WahCUB_200_2011} dataset.
As shown in~\figref{qualitative_cub}, our algorithm shows promise for extending to non-rigid objects as well and warrants further investigation along this direction.

\begin{figure}
	\centering
    \includegraphics[width=0.85\linewidth]{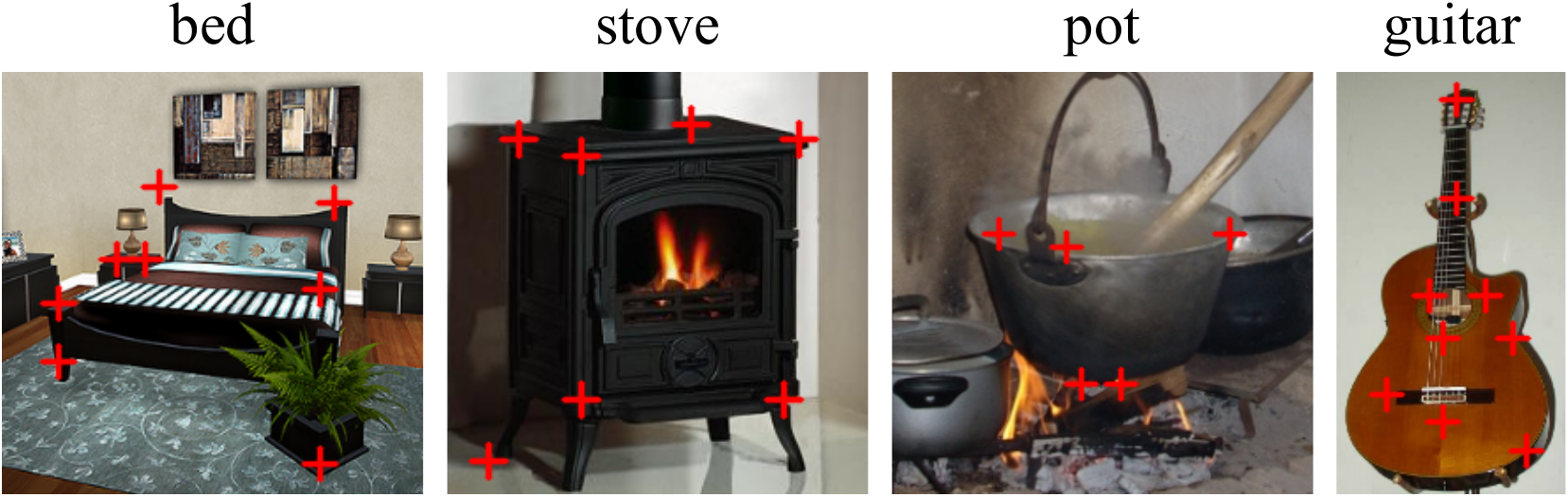}
    \caption{\textbf{Multi-peak heatmap results.} The category-agnostic feature extraction module computed general visible keypoints which helps the viewpoint estimation module to recognize the target object.}
    \label{figure:priormap}
 \end{figure}

\begin{figure}
	\centering
    \includegraphics[width=0.95\linewidth]{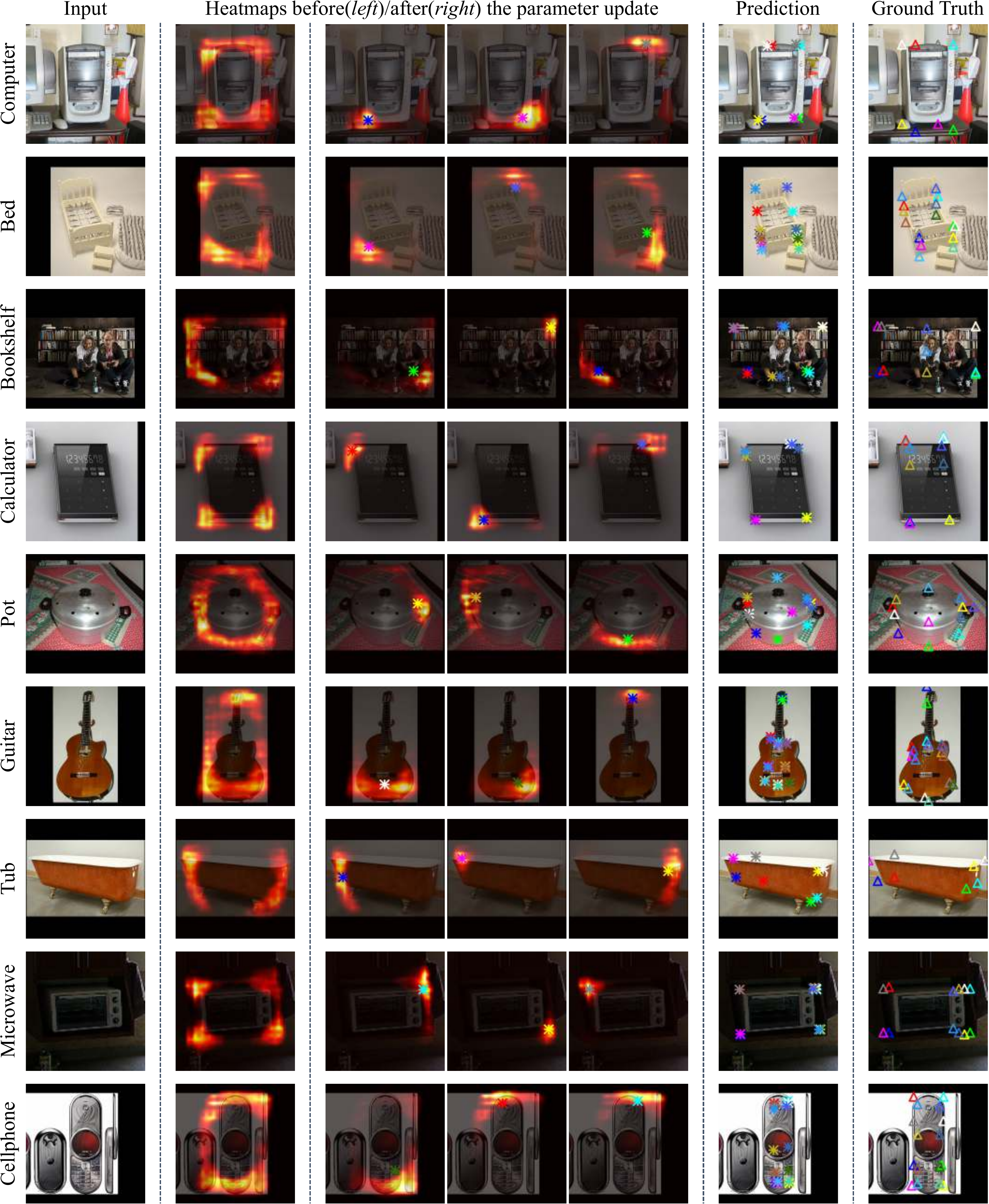}
    \caption{\textbf{Qualitative results of the intra-dataset experiment.} We show more keypoint detection results of MetaView on query images, before and after optimizing it with $10$ support images of a novel category. The images from left to right are: the input image, the 2D keypoint heatmap before the network's parameters are updated for a particular category, three example heatmaps for specific keypoints after the parameter update, the predicted keypoints, and the ground truth keypoints.}
    \label{figure:qualitative_supp}
 \end{figure}

\Paragraph{More Qualitative results.}
We show additional quantitative results for the intra-dataset experiment, including more categories of objects in~\figref{qualitative_supp}. It can be observed that our proposed MetaView algorithm is able to quickly learn to detect category-specific semantic keypoints for viewpoint estimation from just a few training examples of a new category.
We also show the results of multi-peak heatmap computed from the category-agnostic feature extraction module. As shown in~\figref{priormap}, the general keypoints heatmap detects many image feature points, which, much like SIFT/SURF \etc, can help the viewpoint estimation module recognize the target object.

\begin{table}[h]
	\caption{\textbf{Detailed configuration of the proposed network.} We use the following abbreviations: N = Number of filters, K = Kernel size, S = Stride size, P = Padding size, D = Dilation size. ``\textit{conv}", ``Conv", and ``BN" denote a convolution block in ResNet-18~\cite{he2016deep}, convolutional layer with ReLU activation, and a batch normalization layer, respectively.}
    \label{tab:arch}
	\centering
    \begin{tabular}[t]{lc}
    \toprule
    Layer & Image feature extractor \\
    \midrule
    1& \textit{conv1}\\
    2& \textit{conv2}\\
    3& \textit{conv3} \\
    4& \textit{conv4d2}\\
    \bottomrule
    \end{tabular}
    \quad
    \begin{tabular}[t]{lc}
    \toprule
    Layer & General keypoints detector \\
    \midrule
    1& \textit{conv1}\\
    2& \textit{conv2}\\
    3& \textit{conv3} \\
    4& \textit{conv4d2}\\
    5& \textit{conv5d4}\\
    6& Conv(N1-K3-S1-P1)\\
    \bottomrule
    \end{tabular}
    \vspace{1mm}
    
    \begin{tabular}[t]{lc}
    \toprule
    Layer & $f_{\theta_{cat}}$ \\
    \midrule
    1& \textit{conv5d4}\\
    \bottomrule
    \end{tabular}
    \quad 
    \begin{tabular}[t]{lc}
    \toprule
    Layer & $f_{\theta_{key}}$ \\
    \midrule
    1& Conv (N5-K3-S1-P1)\\
    \bottomrule
    \end{tabular}
    \vspace{1mm}
    
    \begin{tabular}[t]{lc}
    \toprule
    Layer & \textit{conv4d2} \\
    \midrule
    1& Conv (N256-K3-S1-P1), BN (N256)\\
    2& Conv (N256-K3-S1-P2-D2), BN (N256)\\
    3& Conv (N256-K3-S1-P2-D2), BN (N256)\\
    4& Conv (N256-K3-S1-P2-D2), BN (N256)\\
    \bottomrule
    \end{tabular}
    \quad 
    \begin{tabular}[t]{lc}
    \toprule
    Layer & \textit{conv5d4} \\
    \midrule
    1& Conv (N512-K3-S1-P1), BN (N512)\\
    2& Conv (N512-K3-S1-P4-D4), BN (N512)\\
    3& Conv (N512-K3-S1-P4-D4), BN (N512)\\
    4& Conv (N512-K3-S1-P4-D4), BN (N512)\\
    \bottomrule
    \end{tabular}
\end{table}

\end{appendices}
\end{document}